\documentclass[conference,a4paper]{IEEEtran}
\IEEEoverridecommandlockouts

\usepackage[hidelinks]{hyperref}
\usepackage[cmex10]{amsmath}
\usepackage{amssymb,amsfonts}
\usepackage{dblfloatfix}

\usepackage[ruled,vlined]{algorithm2e}
\usepackage{graphicx}
\graphicspath{{Figures/PDF/}{Figures/PNG/}}
\usepackage[caption=false,font=footnotesize]{subfig}

\usepackage{booktabs}
\usepackage{siunitx}
\usepackage[numbers,compress]{natbib}
\usepackage{texnames}
\usepackage{bm,bbm}
\usepackage{orcidlink}
\usepackage{multirow}

\title{\uppercase{MANGO: A Global Single-Date Paired Dataset for Mangrove Segmentation}}

\author{\IEEEauthorblockN{Junhyuk Heo, \;\;\; Beomkyu Choi, \;\;\; Hyunjin Shin, \;\;\; Darongsae Kwon}
\IEEEauthorblockA{
TelePIX \\
07330, Seoul, South Korea
\\
\{hjh1037, bkchoi, hyunjin, darong.kwon\}@telepix.net}
}

\begin{document}

\maketitle

\begin{abstract}
Mangroves are critical for climate-change mitigation, requiring reliable monitoring for effective conservation. While deep learning has advanced mangrove detection, its progress is hindered by existing datasets that lack curated single-date image-mask pairs, cover limited regions, or are inaccessible to the public. To address these challenges, we introduce MANGO, a large-scale global dataset of 42,703 labeled image-mask pairs across 124 countries. MANGO is constructed through a general selection framework that automatically pairs each annual mangrove label with the most representative single-date Sentinel-2 acquisition by ranking candidates. We further provide a benchmark across diverse semantic segmentation architectures, establishing a foundation for scalable and reliable global mangrove monitoring. The dataset and code is available at \href{ https://github.com/ROKMC1250/MANGO}{ https://github.com/ROKMC1250/MANGO}
\end{abstract}


\section{Introduction}
\label{sec:intro}
Mangrove forests are a critical ``blue carbon'' ecosystem, storing large amounts of carbon in both biomass and soils while providing shoreline protection and habitat support~\cite{donato2011mangroves, wang2019review}. Despite occupying a relatively small coastal footprint, their loss translates into outsized impacts on carbon budgets and coastal resilience~\cite{donato2011mangroves, wang2019review}.

\begin{figure}[t]
    \centering
    \subfloat[S$2$ ($t_1$)]{\includegraphics[width=0.24\columnwidth]{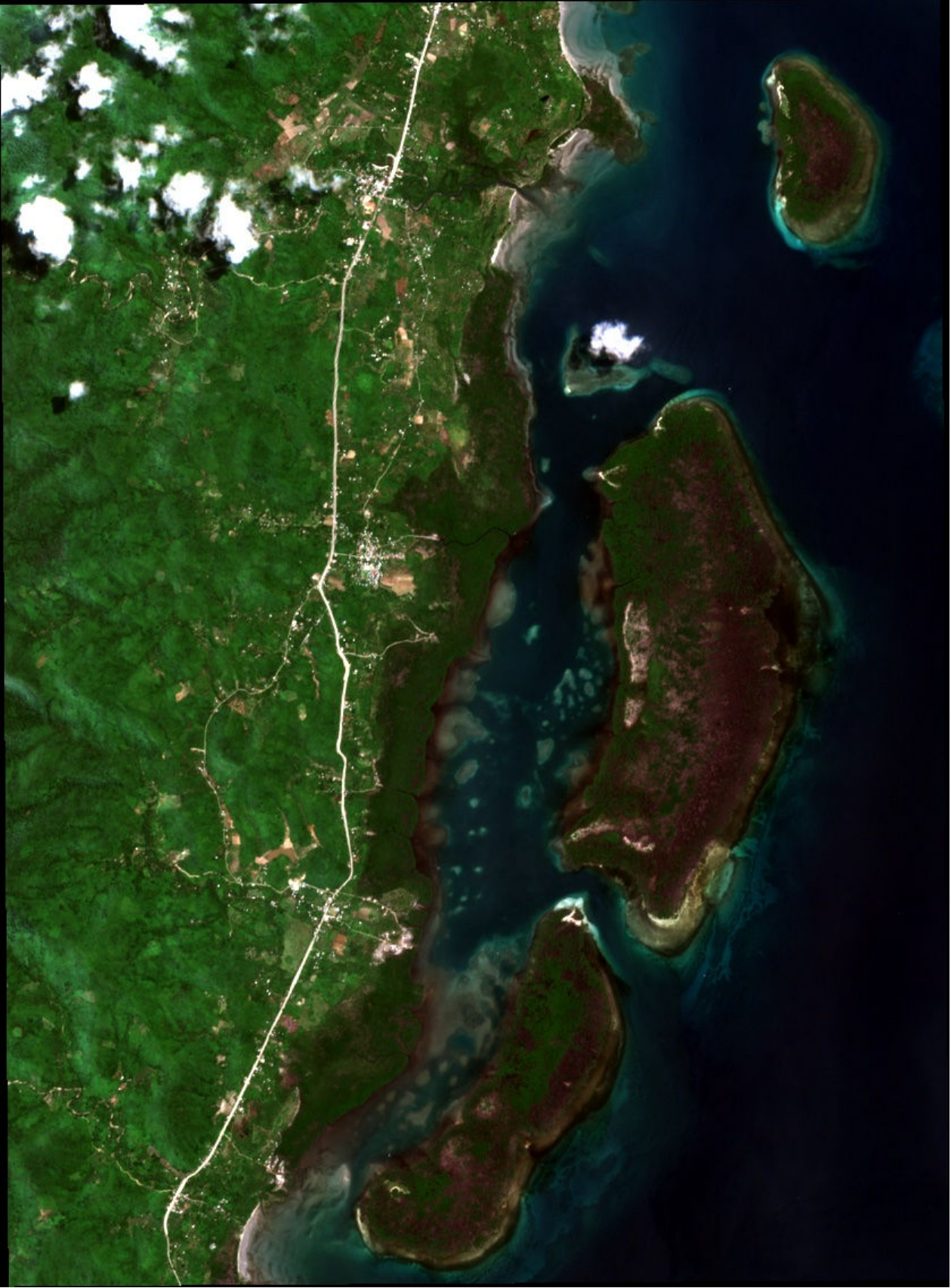}\label{fig:image_a}}
    \hfill
    \subfloat[S$2$ ($t_2$)]{\includegraphics[width=0.24\columnwidth]{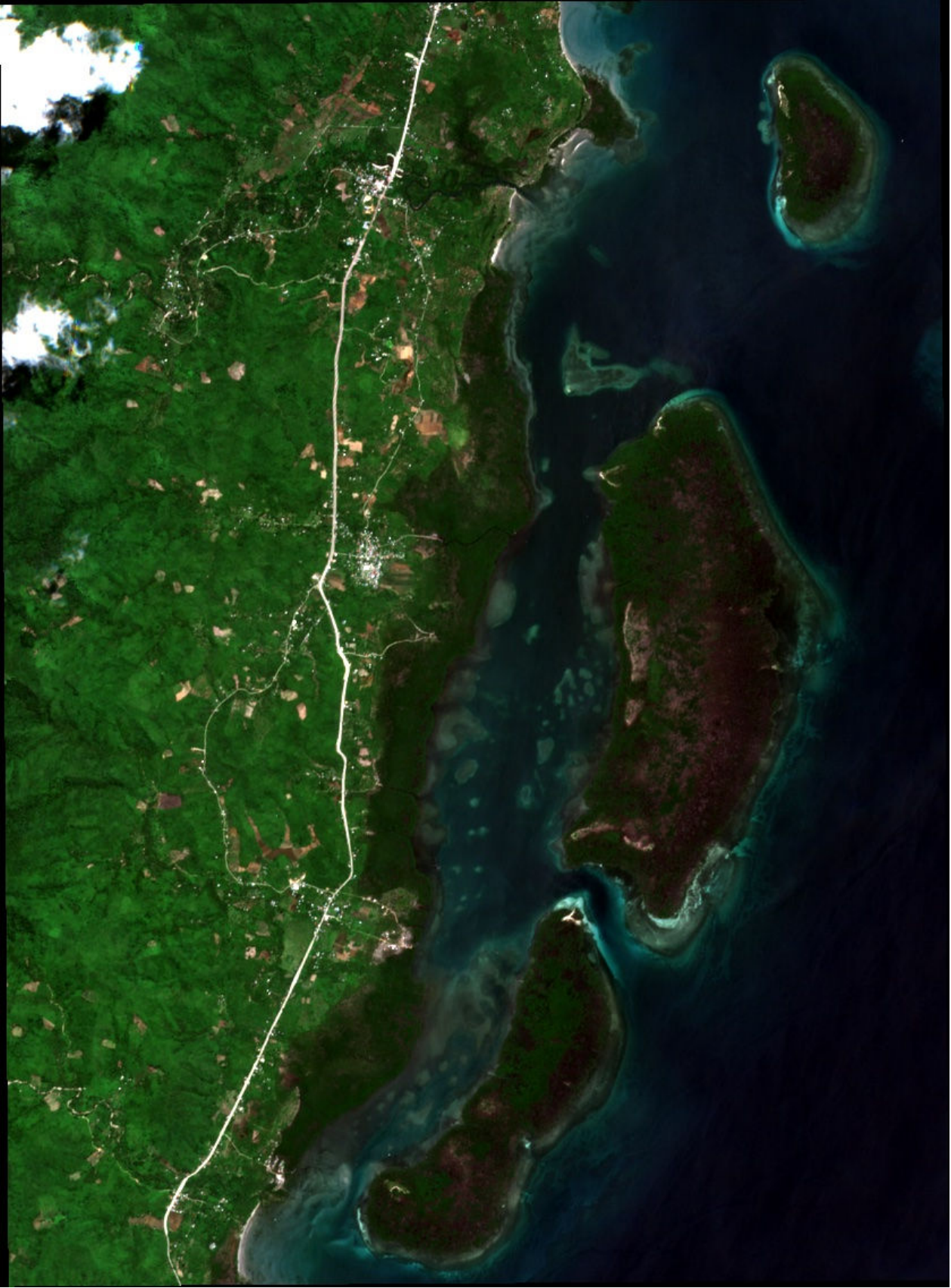}\label{fig:image_c}}
    \hfill
    \subfloat[MVI ($t_1$)]{\includegraphics[width=0.24\columnwidth]{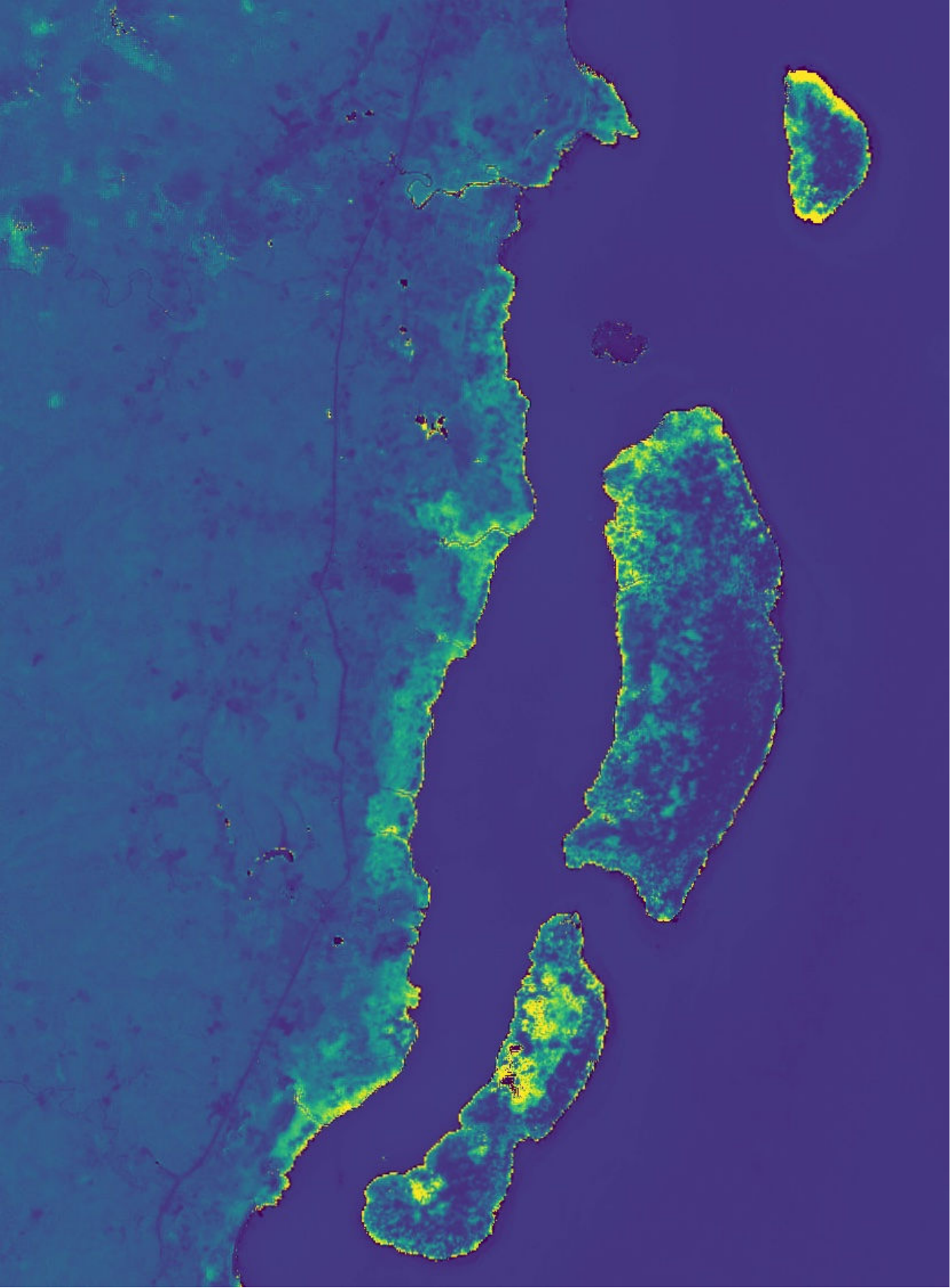}\label{fig:image_b}}
    \hfill
    \subfloat[MVI ($t_2$)]{\includegraphics[width=0.24\columnwidth]{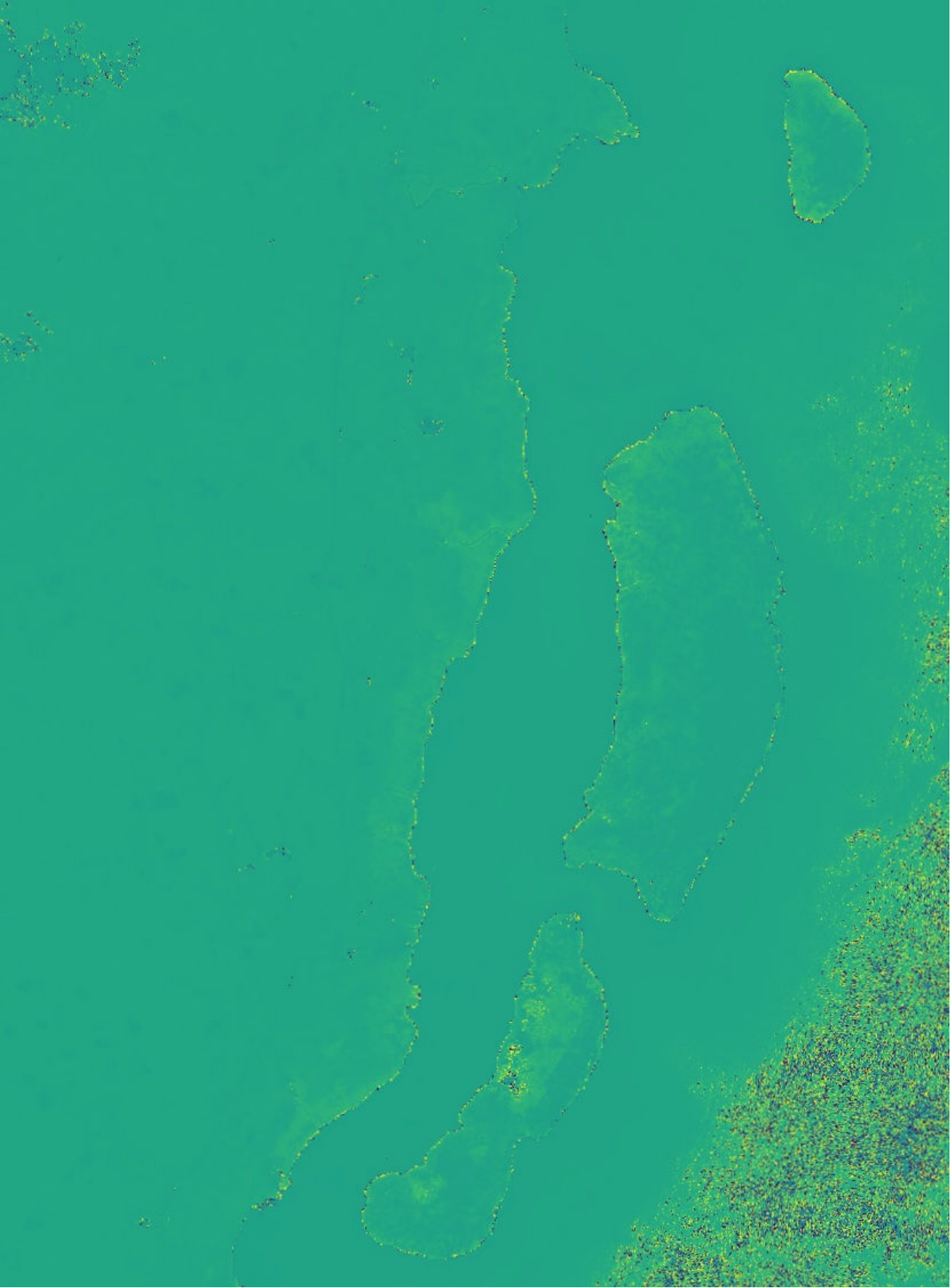}\label{fig:image_d}}
    \caption{Sentinel-2 image pairs and their corresponding MVI responses. 
    (a,b) RGB composites of Sentinel-2 images acquired over the same site at two dates, $t_1$ and $t_2$. 
    (c,d) MVI response maps computed from the corresponding Sentinel-2 images in (a,b), respectively.}
    \label{fig:intro_palawan}
\end{figure}

For decades, remote sensing has been the primary modality for large-scale mangrove monitoring, with operational pipelines relying on spectral analysis such as the Normalized Difference Vegetation Index (NDVI) and Mangrove Vegetation Index (MVI), followed by thresholding~\cite{baloloy2020mvi, tran2022spectral, wang2019review, rouse1974erts}. While attractive due to their simplicity, these approaches suffer from two structural limitations: decision rules require explicit thresholds sensitive to acquisition conditions, and these thresholds rarely transfer across diverse coastlines due to spectral confounders like sediment and mixed pixels~\cite{tran2022spectral, neri2021mvi}. These limitations are evident in Fig.~\ref{fig:intro_palawan}, where two Sentinel-2 Level-2A (S2 L2A) acquisitions over the same site appear visually similar but yield markedly different MVI responses, making threshold-based rules unstable. Motivated by these limitations, the field has shifted toward deep learning, which leverages spatial and spectral patterns rather than fixed thresholds~\cite{iovan2020dcnn, hicks2020hybridcnn}. By learning hierarchical features directly from data, these models generalize across heterogeneous coastlines that previously required site-specific threshold tuning, and subsequent works have continued to expand this paradigm~\cite{guo2021menet, dong2024mangroveseg, zhang2024mwsam, kathiroli2025spatiotemporal}.

\begin{table*}[t!]
\centering
\caption{Comparison of MANGO with existing products and datasets. Availability denotes public access to raw image--label pairs; $\dagger$ indicates repositories that are reported as available but are not practically accessible.}
\label{tab:comparison}
\resizebox{\textwidth}{!}{
\begin{tabular}{lcccccccc}
\hline
Dataset & Scope & Res. (m) & Sensor & 1:1 Pair & Samples & Patch Size & Split Strategy & Availability \\ \hline
GMW v3.0~\cite{bunting2022gmw3} & Global & 25 & SAR / Landsat & No & N/A & N/A & N/A & Map Only \\
HGMF\_2020~\cite{jia2023global10m} & Global & 10 & Sentinel-2 & No & N/A & N/A & N/A & Map Only \\
ME-Net~\cite{guo2021menet} & Regional & 10 & Sentinel-2 & Yes & 5,120 & 256$\times$256 & Random & Restricted \\
TCCFNet~\cite{fu2025tccfnet} & Regional & 0.3 & VHR & Yes & 1,024 & 768$\times$768 & Random & Restricted \\
MagSet-2~\cite{de2024deep} & Global & 10 & Sentinel-2 & Yes & ~10,000+ & 128$\times$128 & Random & Restricted$^\dagger$ \\
MANGO (Ours) & Global & 10 & Sentinel-2 & Yes & 42,703 & 256$\times$256 & Country-disjoint & Public \\ \hline
\end{tabular}
}
\end{table*}

\begin{figure*}[t!]
  \centering
  \includegraphics[width=1.0\textwidth]{Figures/Constructing_MANGO/IGARSS_MANGO.pdf}
  \caption{Scene selection pipeline for constructing single-date image-mask pairs from annual mangrove labels. For each site, multiple Sentinel-2 candidates $\{I_{i,t}\}$ share the same annual mask $M_i$. We extract mangrove reference pixels from $M_i$ to form a target spectrum and compute a detection map $D_{i,t}$. Each candidate is scored by the Fisher discriminant ratio $J(I_{i,t})$ over mangrove and background regions, and the final acquisition is selected by $t^*=\arg\max_t J(I_{i,t})$.}
  \label{fig:overall_pipeline}
\end{figure*}

However, the progress of deep learning approaches has been constrained by how training data have been constructed. Early efforts relied on manual visual interpretation over single sites~\cite{guo2021menet, fu2025tccfnet}, which proved costly and regionally bounded. Recent works have increasingly turned to Global Mangrove Watch (GMW)~\cite{bunting2022gmw3} as a reference label, which together with HGMF~\cite{jia2023global10m} has emerged as a de facto standard for global mangrove extent. MagSet-2~\cite{de2024deep} represents the first global attempt to directly use GMW as supervision, but it relies on temporal composites and a random split that overlooks spatial autocorrelation across nearby sites~\cite{ploton2020spatial}. Such composites blur acquisitions into a synthetic image that does not faithfully represent any real observation, and they mismatch operational monitoring, which is performed on single-date acquisitions.

This trajectory reveals an unresolved problem: GMW provides only annual labels, while supervised segmentation requires single-date imagery aligned with these labels. Prior works have circumvented this mismatch through manual curation, regional restriction, or temporal compositing, leaving open how to automatically pair an annual label with the most representative single-date observation. We define this as the ``temporal pairing gap.''

To bridge this gap, we introduce MANGO, a large-scale global dataset comprising 42,703 curated image-mask pairs across 124 countries. MANGO is constructed through a two-stage pipeline: data collection via Google Earth Engine~\cite{gorelick2017gee}, followed by a general selection framework that automatically identifies the most representative single-date acquisition for each site by ranking candidates with a class-separability criterion. Furthermore, we establish a standardized benchmark under a country-disjoint split, where no country appears in both training and test sets, ensuring rigorous evaluation of geographic generalization. We evaluate various architectures, including CNN-based models~\cite{zhou2018unetplusplus, fan2020manet, li2018pan} and Transformer-based models~\cite{xie2021segformer, lin2017fpn, ranftl2021dpt, xiao2018upernet} under this protocol. A summary comparison with existing mangrove datasets is provided in Table~\ref{tab:comparison}. The primary contributions are as follows:

\begin{itemize}
    \item \textbf{MANGO Dataset}: A public, global, large-scale mangrove segmentation dataset with 42,703 single-date Sentinel-2 L2A images paired with GMW labels, covering 124 countries.
    \item \textbf{Pairing Framework}: A general framework that resolves the temporal pairing gap by automating the alignment of annual labels with single-date observations, supporting multiple interchangeable selection criteria.
    \item \textbf{Benchmark}: A country-disjoint benchmark evaluating diverse segmentation backbones, with selection-criterion ablation establishing principled scene selection over naive baselines.
\end{itemize}


\begin{figure*}[t!]
    \centering
    \subfloat[Local sampling and stratification]{%
        \includegraphics[width=0.29\textwidth]{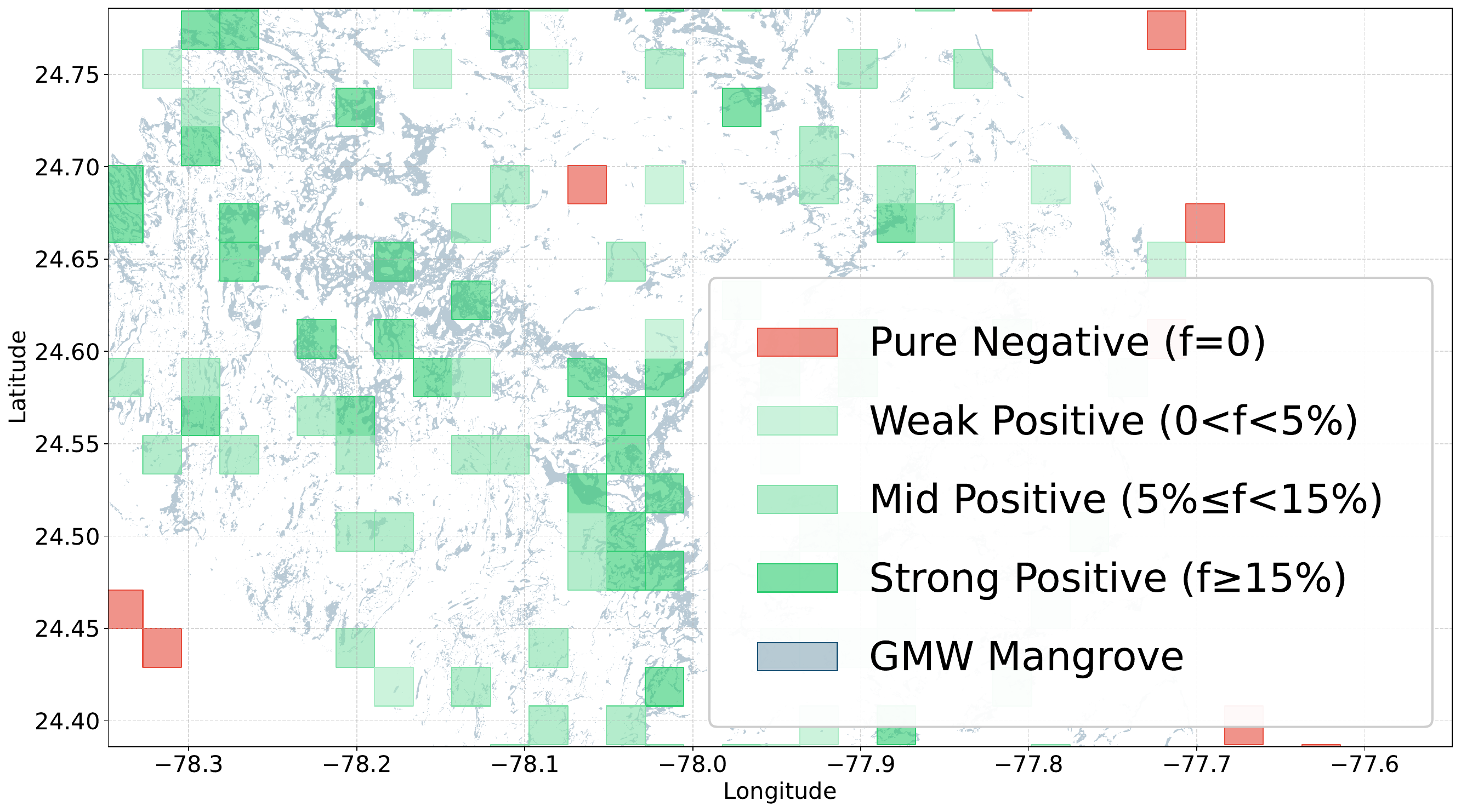}%
        \label{fig:local_sampling}%
    }\hfill
    \subfloat[Global sampling footprint]{%
        \includegraphics[width=0.33\textwidth]{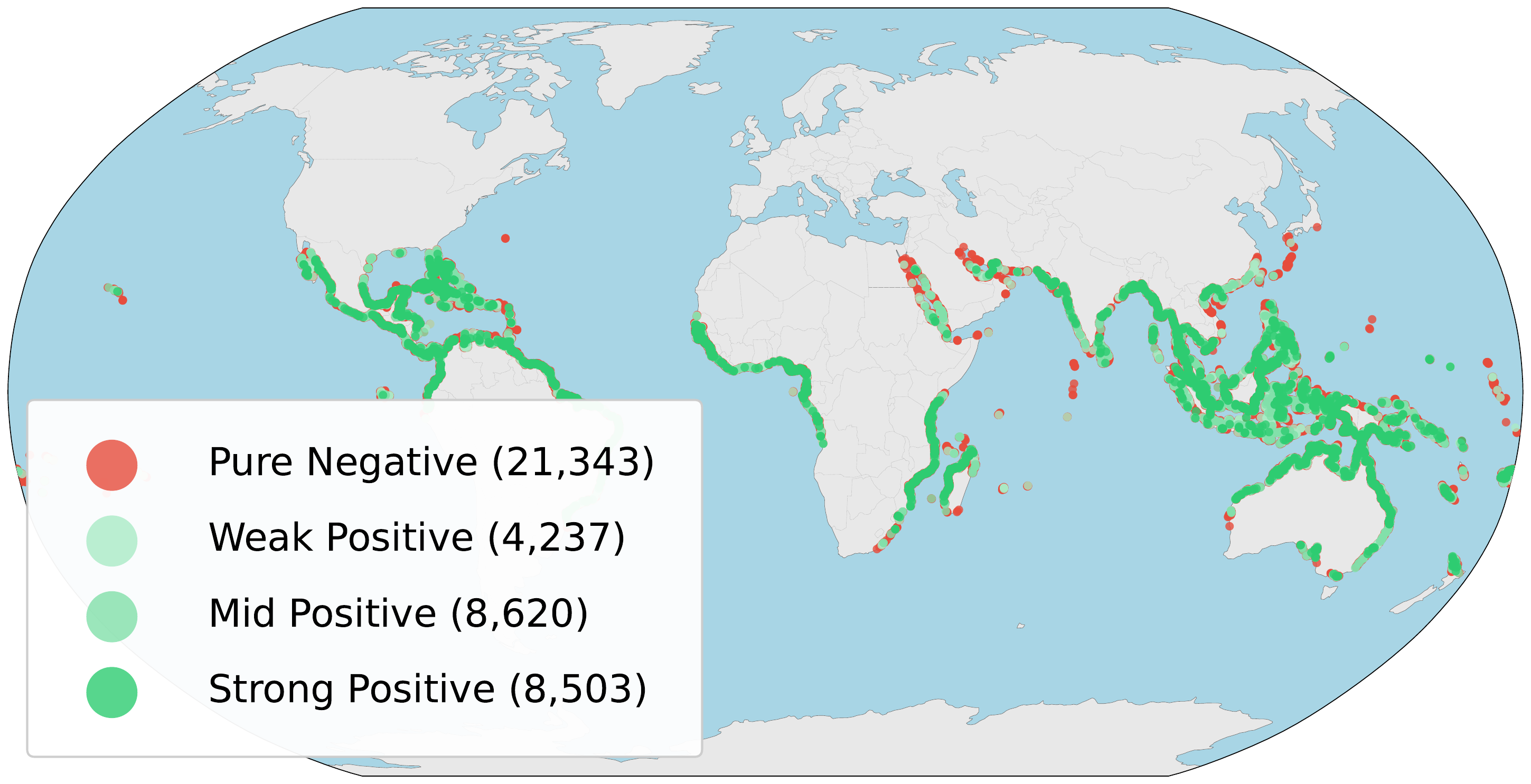}%
        \label{fig:footprint}%
    }\hfill
    \subfloat[Country-disjoint data split]{%
        \includegraphics[width=0.33\textwidth]{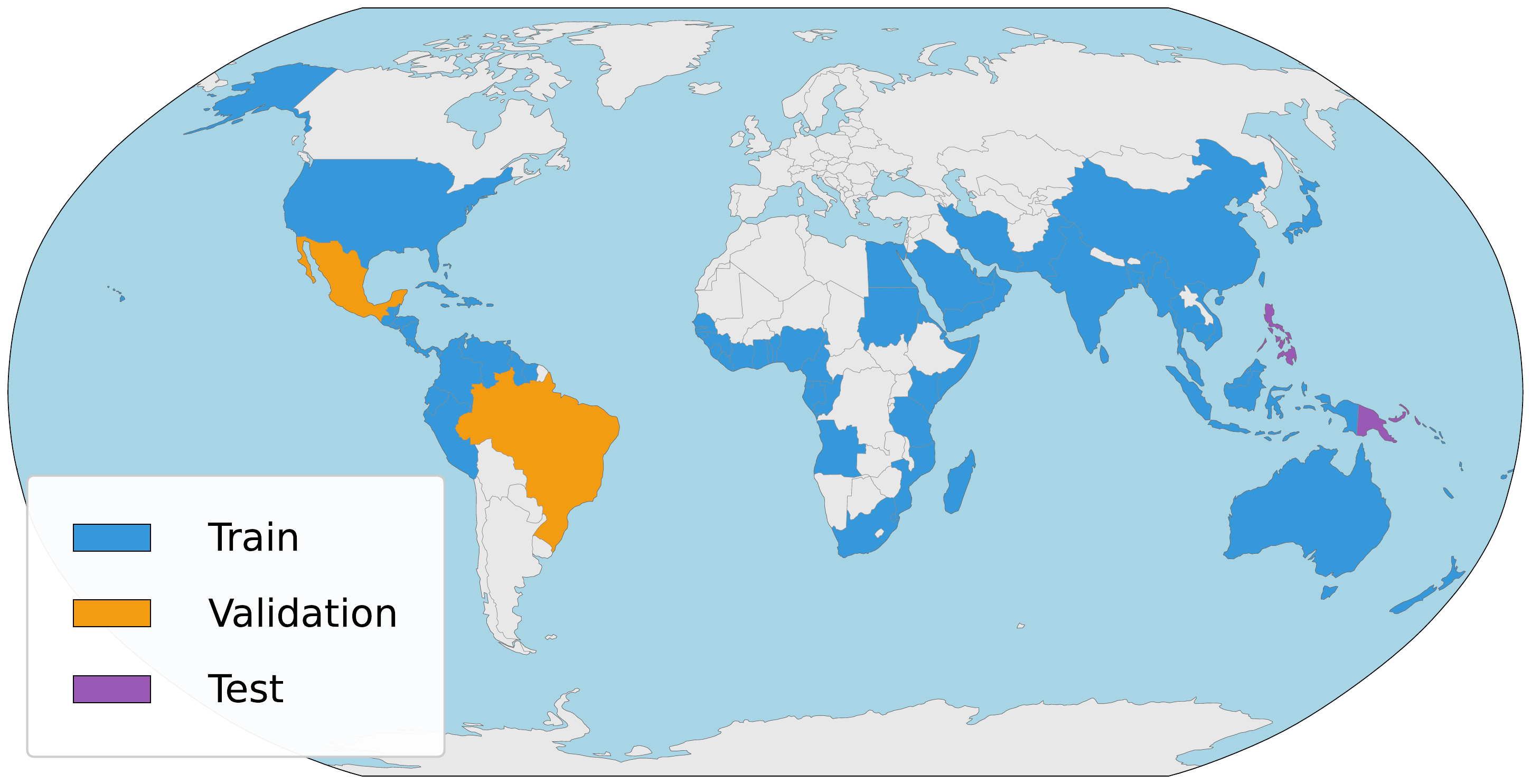}%
        \label{fig:split}%
    }
    \caption{Global overview and experimental setup of the MANGO dataset. (a) illustrates the categorization of tiles into positive and negative classes. (b) shows the distribution of MANGO images across diverse countries. (c) visualizes the geographic partition used for rigorous generalization testing.}
    \label{fig:dataset_overview}
\end{figure*}

\section{Constructing MANGO}
We construct MANGO through a scalable pipeline that resolves the temporal pairing gap by automatically aligning each annual GMW label with the most representative single-date Sentinel-2 acquisition. As illustrated in Fig.~\ref{fig:overall_pipeline}, the pipeline consists of large-scale data collection via GEE~\cite{gorelick2017gee} followed by a selection framework that admits multiple interchangeable score functions.

\subsection{Dataset Collection}
We tile the GMW~\cite{bunting2022gmw3} extent into a set of regions $\mathcal{R}=\{R_i\}$, where each $R_i$ is a $256 \times 256$ pixel patch at 10\,m resolution~\cite{bunting2022gmw3}. For each sensing date $t$, we denote the corresponding Sentinel-2 L2A acquisition over $R_i$ by $I_{i,t}\in\mathbb{R}^{256\times256\times 13}$. For each $R_i$, we define the candidate pool $\mathcal{I}_i$ of S2 L2A~\cite{drusch2012sentinel2} acquisitions $I_{i,t}$ as follows:
\begin{equation}
\mathcal{I}_i = \{ I_{i,t} \mid t \in \mathcal{T}_i, \, C(I_{i,t}) < \kappa, \, \Omega(I_{i,t}) \ge \omega \},
\end{equation}
where $\mathcal{T}_i$ denotes the sensing dates within the target year 2020~\cite{bunting2022gmw3}. To ensure data quality, we define $C(I_{i,t})$ as the cloud fraction calculated using the cloud mask provided by GEE~\cite{gorelick2017gee}, and $\Omega(I_{i,t})$ as the spatial coverage of the area of interest within the region. We set the strict filtering thresholds to $\kappa=0.05$ and $\omega=0.50$. In total, this pipeline retrieved 1,379,743 images across 42,703 regions, providing an average of 32 candidates per site. The central problem addressed in the next stage is the automatic selection of one representative acquisition from this pool.

\subsection{Selection Framework}
Given a candidate pool $\mathcal{I}_i$ and the annual mask $M_i$, we cast scene selection as a ranking problem driven by class separability. The framework consists of two components: a score function that produces a per-pixel score map for each candidate, and a separability criterion that ranks candidates by how distinctly the score map aligns with the annual label.

\noindent\textbf{Score function.} A score function is a mapping $f:\mathbb{R}^{H\times W\times B}\to\mathbb{R}^{H\times W}$ that assigns to each pixel a continuous score reflecting its likelihood of belonging to the mangrove class. For each $I_{i,t}\in\mathcal{I}_i$, $f$ produces a per-pixel score map $D_{i,t}=f(I_{i,t})$. We do not restrict the functional form of $f$: it may be a fixed spectral index, a scene-adaptive detector, or any other pixel-wise scoring rule. A useful $f$ should yield separable score distributions over mangrove and background regions, so that the date with the strongest contrast can be identified by a class-separability criterion.

\noindent\textbf{Ranking by class separability.} Given $D_{i,t}$, we evaluate how distinctly it separates the two regions defined by the annual mask $M_i$. Let $\Omega_{m,i}=\{\mathbf{x}|M_i(\mathbf{x})=1\}$ and $\Omega_{b,i}=\{\mathbf{x}|M_i(\mathbf{x})=0\}$ denote the mangrove and background sets, with class-wise mean and variance
\begin{equation}
\mu_{c,t}=\text{mean}(D_{i,t}(\Omega_{c,i})),\;\;\sigma_{c,t}^2=\text{var}(D_{i,t}(\Omega_{c,i})),
\end{equation}
where $c\in\{m,b\}$. Class separability is quantified by the Fisher discriminant ratio (FDR):
\begin{equation}
J(I_{i,t})=\frac{(\mu_{m,t}-\mu_{b,t})^2}{\sigma_{m,t}^2+\sigma_{b,t}^2}.
\end{equation}
A higher $J$ indicates that the score map more cleanly separates label-defined mangrove and background regions, providing more reliable supervision for downstream segmentation. The optimal acquisition for region $i$ is then selected as $t^*=\arg\max_{t\in\mathcal{T}_i} J(I_{i,t})$. This formulation is agnostic to $f$, allowing different score functions to be plugged into the same evaluation protocol.

\noindent\textbf{Score function instantiations.} We consider two complementary score functions within this framework. The first is a fixed spectral index, the MVI~\cite{baloloy2020mvi}, which computes per-pixel scores from a closed-form combination of spectral bands without scene-specific parameters. The second is a scene-adaptive target detector. Among target detection methods, we adopt the matched filter (MF)~\cite{fuhrmann1992cfar} as a representative instantiation. We extract a target spectrum $\mathbf{s}_{i,t}\in\mathbb{R}^{13}$ from high-purity pixels of $M_i$, and estimate the background mean $\boldsymbol{\mu}_{i,t}$ and covariance $\Gamma_{i,t}$ from pixels outside $M_i$. The score map is then
\begin{equation}
D_{i,t}(\mathbf{x})=\frac{\mathbf{s}_{i,t}^{\top}\Gamma_{i,t}^{-1}(\mathbf{x}-\boldsymbol{\mu}_{i,t})}{\sqrt{\mathbf{s}_{i,t}^{\top}\Gamma_{i,t}^{-1}\mathbf{s}_{i,t}}}.
\end{equation}
Intuitively, the score measures how closely each pixel resembles the scene-specific mangrove signature after suppressing background variability through whitening. While MVI relies on a globally fixed formulation, the target-detection score adapts to per-scene background statistics, suppressing site-specific confounders such as sediment plumes and turbid water. Both instantiations are valid within our framework, and we evaluate them as alternative selection protocols in Section~\ref{sec:experiments}. By decoupling scene selection into a score function and a class-separability criterion, the framework provides a standardized testbed for comparing existing or future score functions.

\begin{table*}[t!]
  \small
  \centering
    \caption{Benchmark results of segmentation models on the MANGO country-disjoint test set under different selection protocols. Random denotes a baseline constructed from randomly selected image--label pairs. Each entry reports IoU/F1 (\%). Formatting: \textbf{bold} marks the best score in each column and \underline{underline} marks the second best.}
  \label{tab:quantitative}
  \renewcommand{\arraystretch}{1.2}
  \begin{tabular}{lccccccc}
    \toprule
    \textbf{Selection} 
    & \textbf{UNet++~\cite{zhou2018unetplusplus}} 
    & \textbf{MAnet~\cite{fan2020manet}} 
    & \textbf{PAN~\cite{li2018pan}} 
    & \textbf{SegFormer~\cite{xie2021segformer}} 
    & \textbf{FPN~\cite{lin2017fpn}} 
    & \textbf{DPT~\cite{ranftl2021dpt}} 
    & \textbf{UPerNet~\cite{xiao2018upernet}} \\
    \midrule
    Random
    & 83.37 / 84.92
    & 82.85 / 83.33
    & 80.41 / 81.83
    & 79.51 / 81.17
    & 75.85 / 77.03
    & 79.98 / 80.34
    & 83.71 / 84.67 \\
    
    MVI-based~\cite{baloloy2020mvi}
    & \underline{87.73 / 89.49}
    & \underline{87.93 / 89.79}
    & \underline{88.92 / 90.72}
    & \textbf{89.95 / 91.73}
    & \textbf{91.10 / 92.66}
    & \textbf{86.15 / 87.55}
    & \underline{88.54 / 90.29} \\
    
    MF-based~\cite{fuhrmann1992cfar}
    & \textbf{91.47 / 93.21}
    & \textbf{91.29 / 93.00}
    & \textbf{90.30 / 92.05}
    & \underline{87.56 / 89.50}
    & \underline{90.70 / 92.48}
    & \underline{84.84 / 86.88}
    & \textbf{88.73 / 90.66} \\
    \bottomrule
  \end{tabular}
\end{table*}

\begin{figure*}[t!]
  \centering
  \subfloat[Input]{%
    \includegraphics[width=0.105\textwidth]{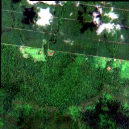}%
  }\hfill
  \subfloat[GT]{%
    \includegraphics[width=0.105\textwidth]{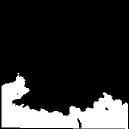}%
  }\hfill
  \subfloat[UNet++]{%
    \includegraphics[width=0.105\textwidth]{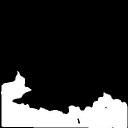}%
  }\hfill
  \subfloat[MAnet]{%
    \includegraphics[width=0.105\textwidth]{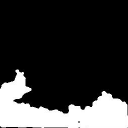}%
  }\hfill
  \subfloat[PAN]{%
    \includegraphics[width=0.105\textwidth]{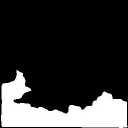}%
  }\hfill
  \subfloat[SegFormer]{%
    \includegraphics[width=0.105\textwidth]{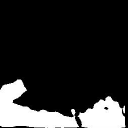}%
  }\hfill
  \subfloat[FPN]{%
    \includegraphics[width=0.105\textwidth]{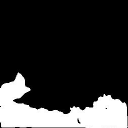}%
  }\hfill
  \subfloat[DPT]{%
    \includegraphics[width=0.105\textwidth]{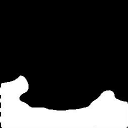}%
  }\hfill
  \subfloat[UPerNet]{%
    \includegraphics[width=0.105\textwidth]{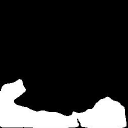}%
  }
  \caption{\textbf{Qualitative results across baseline models.} (a) Sentinel-2 L2A imagery, (b) MANGO ground-truth mask established via the quality-aware selection pipeline, and (c--i) segmentation predictions from baseline models.}
  \label{fig:qualitative}
\end{figure*}

\subsection{Dataset Information}
The MANGO dataset consists of 42,703 curated 13-band image-mask pairs at 10\,m resolution, spanning 124 countries. As shown in Fig.~\ref{fig:footprint}, these samples are distributed worldwide to capture diverse coastal environments, with a composition of 21,343 Pure Negative, 4,237 Weak Positive, 8,620 Mid Positive, and 8,503 Strong Positive images. These categories are defined based on the mangrove fraction within each tile relative to the GMW annual label as illustrated in Fig.~\ref{fig:local_sampling}. While prior mangrove datasets typically adopt random splits~\cite{guo2021menet, fu2025tccfnet, de2024deep}, such splits are known to overestimate generalization in geospatial settings due to spatial autocorrelation between nearby samples~\cite{ploton2020spatial}. To enable rigorous evaluation that reflects deployment across unseen coastlines, we adopt a country-disjoint split protocol with an 8:1:1 ratio for training, validation, and testing (Fig.~\ref{fig:split}), ensuring that no country appears in both training and test sets.

\section{Experiments}
\label{sec:experiments}
All experiments were implemented in PyTorch using NVIDIA RTX 4090 GPUs. Models were optimized via AdamW with $10^{-3}$ learning rate for 50 epochs under identical hyperparameter configurations. All evaluations are conducted on the MANGO country-disjoint split.

\subsection{Benchmark Results}
We benchmark seven segmentation architectures on datasets constructed under three selection protocols: random selection as a control baseline, and the two principled instantiations of our framework, MVI~\cite{baloloy2020mvi} and MF~\cite{fuhrmann1992cfar}. 
The Random protocol draws one acquisition per site uniformly from the same filtered candidate pool $\mathcal{I}_i$ defined in Eq.~(1), without applying any score function or class-separability criterion. It therefore shares the cloud and coverage filtering with the other two protocols and isolates the contribution of score-guided ranking itself, rather than the effect of basic data quality control.
As shown in Table~\ref{tab:quantitative}, both MVI- and MF-based selection consistently outperform Random across all architectures, confirming that score-guided scene selection provides measurably stronger supervision than naive sampling. 
Random remains lower than the two principled protocols across all backbones.
The relative ranking between MVI and MF varies across architectures, with neither uniformly dominating, indicating that the framework is robust to the choice of score function rather than being tied to a single optimal detector.

\begin{figure}[t!]
\centering
\renewcommand{\arraystretch}{0.5}
\setlength{\tabcolsep}{1pt}

\begin{tabular}{@{}c@{\hspace{1.5mm}} cccc@{}}
& RGB & Mask & MVI & MF \\[1.5mm]

\rotatebox{90}{\parbox[c]{2cm}{\centering S2 ($t_1$)}} &
\includegraphics[width=0.23\columnwidth]{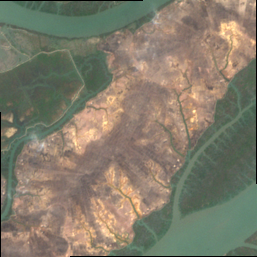} &
\includegraphics[width=0.23\columnwidth]{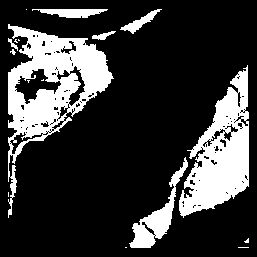} &
\includegraphics[width=0.23\columnwidth]{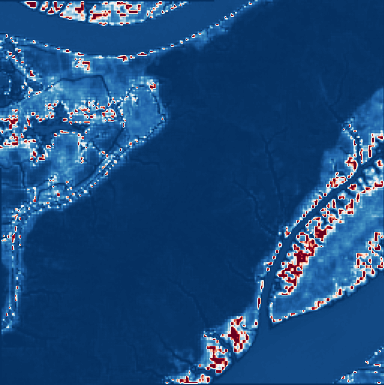} &
\includegraphics[width=0.23\columnwidth]{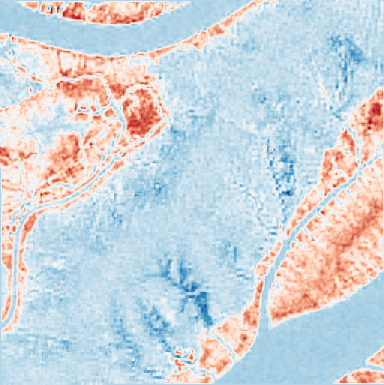} \\
&  &  & (0.46) & \textbf{(2.87)} \\[1mm]

\rotatebox{90}{\parbox[c]{2cm}{\centering S2 ($t_2$)}} &
\includegraphics[width=0.23\columnwidth]{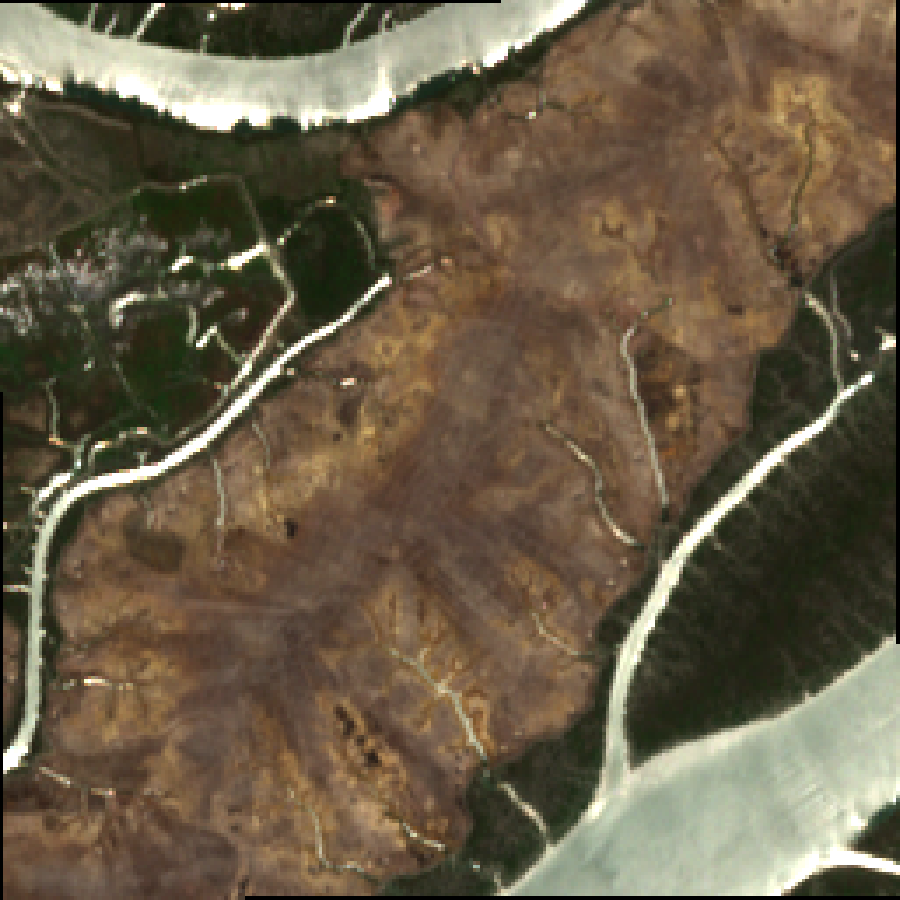} &
\includegraphics[width=0.23\columnwidth]{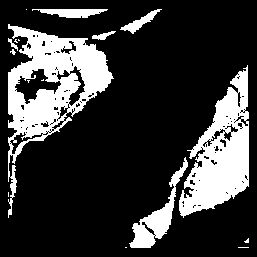} &
\includegraphics[width=0.23\columnwidth]{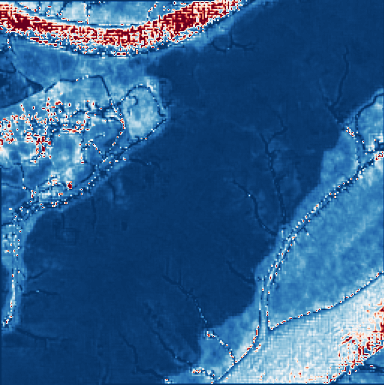} &
\includegraphics[width=0.23\columnwidth]{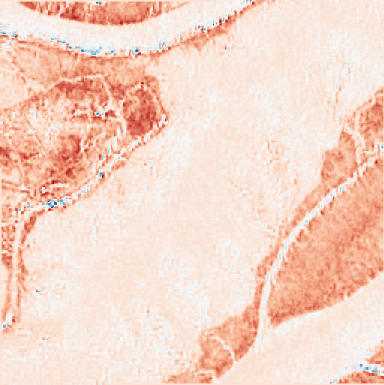} \\
&  &  & (0.11) & \textbf{(1.92)} \\[1mm]

\end{tabular}
\caption{Framework operation under two score function instantiations for two Sentinel-2 candidates of the same region. The mask is the shared GMW-derived annual label. MVI and MF produce per-pixel score maps ranked by the same FDR protocol, with values in parentheses reporting $J(I_{i,t})$.}

\label{fig:selection_comparison}
\end{figure}

\subsection{Qualitative Analysis}
Figure~\ref{fig:qualitative} shows segmentation predictions across architectures on a representative scene from the country-disjoint test set. Models trained on MANGO delineate sharp coastal boundaries and retain small, fragmented mangrove stands, with cleaner outputs in turbid and sediment-rich regions. Figure~\ref{fig:selection_comparison} further illustrates how the framework operates under MVI and MF score functions: both produce per-pixel score maps ranked by the same FDR protocol, with characteristic responses reflecting their structural differences. These results indicate that the supervision provided by our pairing framework transfers effectively across diverse global coastal contexts.

\section{Conclusion}
We presented MANGO, a public global dataset of 42,703 single-date Sentinel-2 image-mask pairs spanning 124 countries, designed to enable supervised mangrove segmentation at scale. To bridge the temporal pairing gap between annual labels and single-date observations, we proposed a general selection framework that ranks candidate acquisitions by class separability and admits multiple interchangeable score functions. Benchmarking under a country-disjoint split shows that principled instantiations of the framework consistently improve over naive selection across diverse architectures, providing cleaner supervision across diverse coastal conditions. The framework currently operates on single-date selections, leaving multi-date fusion as a natural extension. We expect MANGO to support reproducible benchmarking and more reliable global mangrove monitoring for conservation and climate-related applications.

\small
\bibliographystyle{IEEEtranN}
\bibliography{references}

\end{document}